\documentclass[runningheads]{llncs}

\usepackage{bbm}
\usepackage{orcidlink}
\usepackage[T1]{fontenc}
\usepackage{overpic}
\usepackage{xcolor}
\usepackage{enumitem}
\usepackage{dirtree}
\usepackage{caption}
\usepackage{multirow}
\usepackage{amsmath,amssymb,amsfonts}
\usepackage{booktabs}
\usepackage{bm}
\usepackage{esvect}
\usepackage{subcaption}

\PassOptionsToPackage{hyphens}{url}

\usepackage{hyperref}
\hypersetup{breaklinks=true,colorlinks=false,pdfborder={0 0 0}}

\usepackage{graphicx}
\begin{document}
\title{Indoor Heat Estimation \\from a Single Visible-Light Panorama}

\author{Guanzhou Ji\orcidlink{0000-0002-0673-3286} 
\and Sriram Narayanan\orcidlink{0000-0002-9303-1791}
\and \\ Azadeh O. Sawyer\orcidlink{0000-0003-1750-5892} 
\and Srinivasa G. Narasimhan\orcidlink{0000-0003-0389-1921}}
\authorrunning{G. Ji et al.}
%
\institute{Carnegie Mellon University, Pittsburgh, USA\\
\email{\{gji,snochurn,asawyer,srinivas\}@andrew.cmu.edu}}

\maketitle              
\begin{abstract}

This paper introduces a novel image-based rendering technique for jointly estimating indoor lighting and thermal conditions from paired indoor-outdoor high dynamic range (HDR) panoramas. Our method uses the indoor panorama to estimate the 3D floor layout, while the corresponding outdoor panorama serves as an environment map to infer spatially-varying illumination and material properties. Assuming indoor surfaces are Lambertian and that all heat originates from outdoor visible light, we model the relationship between light transport and heat transfer, and perform transient heat simulations to generate indoor temperature distributions. The simulated heat maps are validated against real-world thermal images captured with an infrared camera. This approach supports photorealistic and physically informed visualization, enabling integrated light and heat estimation to advance traditional virtual home staging.

\end{abstract}
\section{Introduction}
Omnidirectional photography enables the capture of a single image that comprehensively represents the entire indoor scene. Recent studies \cite{zhi2022semantically,ji2023virtual,ji2024virtual,ji2025digital} have shown that a single indoor panorama can be used to estimate spatially-varying light, material properties, and 3D floor layouts, supporting applications such as indoor virtual staging. As outdoor light enters a space, it not only shapes the visual appearance of the interior, but it also introduces solar radiation that contributes to heat accumulation on indoor surfaces. Estimating the heat distribution is essential for evaluating the potential heating and cooling demands throughout the space. However, thermal assessment typically relies on infrared thermal cameras, which are expensive, prone to noise, and limited to capturing 2D perspective images with narrow field of view (FOV). These limitations hinder the ability to obtain high-quality thermal information across an entire 360$^\circ$ indoor scene. This challenge motivates a key research question: \textit{Can we directly estimate indoor thermal properties from a single visible-light panorama?}

\begin{figure*}
  \centering
  \begin{overpic}[width= \textwidth]{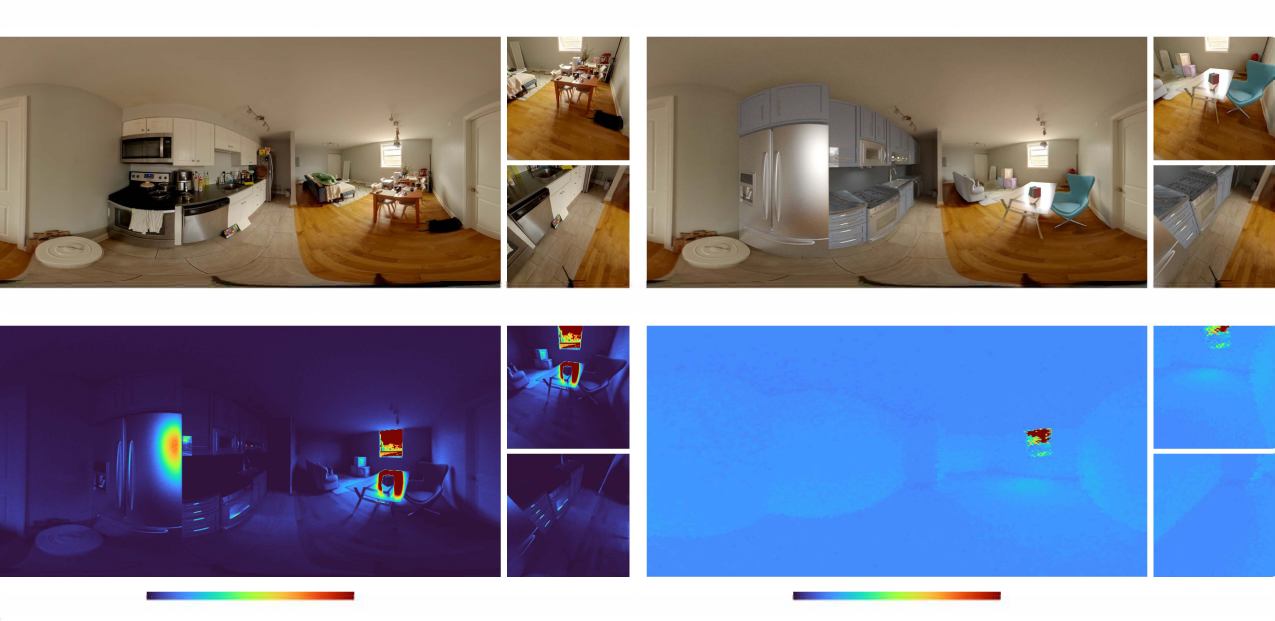}
    \put(0,47){\tiny \color{black}{(a) Captured Scene}}
    \put(51,47){\tiny \color{black}{(b) Edited Scene}}
    \put(0,24){\tiny \color{black}{(c) Light Estimation}}
    \put(51,24){\tiny \color{black}{(d) Heat Estimation}} 
    \put(10,1.5){\tiny \color{black}{0}}
    \put(28.5,1.5){\tiny \color{black}{3,000$cd/m^2$}}
    \put(57,1.5){\tiny \color{black}{21$^{\circ}\mathrm{C}$}}
    \put(79,1.5){\tiny \color{black}{25$^{\circ}\mathrm{C}$}} 
  \end{overpic}
  \caption{Overview of this work: An indoor panorama (a) is captured under natural illumination, with a paired outdoor panorama providing real-time, spatially-varying light to the scene. The scene is virtually edited with new indoor layout objects (b). The absolute light level ($cd/m^2$) (c) is estimated for the virtual scene, and we introduced a heat transport equation to compute an indoor heat map (d) displaying per-pixel temperature values ($^{\circ}\mathrm{C}$).}
  \label{fig_ht_teaser}
\end{figure*}

Real-world lighting varies over time with spectral characteristics. To model outdoor illumination, virtual sky models \cite{cie2003spitial,perez1993all,hosek2012analytic} have been developed for light estimation. The global rendering method \cite{debevec2006image} provides a high dynamic range (HDR) image-based rendering setup for relighting virtual objects within realistic scene contexts. More recently, image-based rendering techniques have used captured or estimated HDR images to achieve instantaneous light estimation from images. HDR techniques capture the appearance of the entire sky, including direct sunlight and atmospheric conditions, allowing an accurate representation of the luminance distribution of the sky and the surrounding urban context \cite{inanici2009applications,inanici2010evalution}. By using such outdoor HDR images as illumination sources, indoor environments can be rendered at the pixel level with realistic lighting effects. 
 
Intrinsic decomposition models light transport by splitting an image into reflectance (albedo) and shading (illumination) components. Intrinsic decomposition has been applied in various tasks, including image editing \cite{bonneel2017intrinsic}, removing interreflections \cite{seitz2005theory}, and separating illumination for material recoloring \cite{carroll2011illumination}, as well as being applied to multi-view outdoor photos \cite{duchene2015multi}. To construct indoor-outdoor light transport, previous studies \cite{ji2023virtual,ji2024virtual} used an indoor panorama and a paired outdoor hemispherical image to reconstruct the indoor appearance. In this work, we simplify the rendering setup using indoor-outdoor panoramas captured with a single camera. This method streamlines data collection and post-processing.

Heat transport includes the process of heat generated by the source and its exchange between different mediums \cite{bergman2011introduction}. Temperature exchange within an object follows the well-known heat transfer physics, and three primary mechanisms include conduction, convection, and radiation. Recent work by Ramanagopal et al. \cite{ramanagopal2024theory} bridges visible light transport and heat transport in solids and estimates heat generation from light absorption. Previous studies on heat estimation have primarily focused on small objects under controlled artificial lighting and captured thermal images in a 2D perspective. In this work, we estimate the absolute energy influx on indoor surfaces using photometrically calibrated indoor–outdoor HDR images and compute the resulting heat generated by outdoor light on these surfaces.

\begin{figure*}[t]
  \centering
  \begin{overpic}[width=\textwidth]{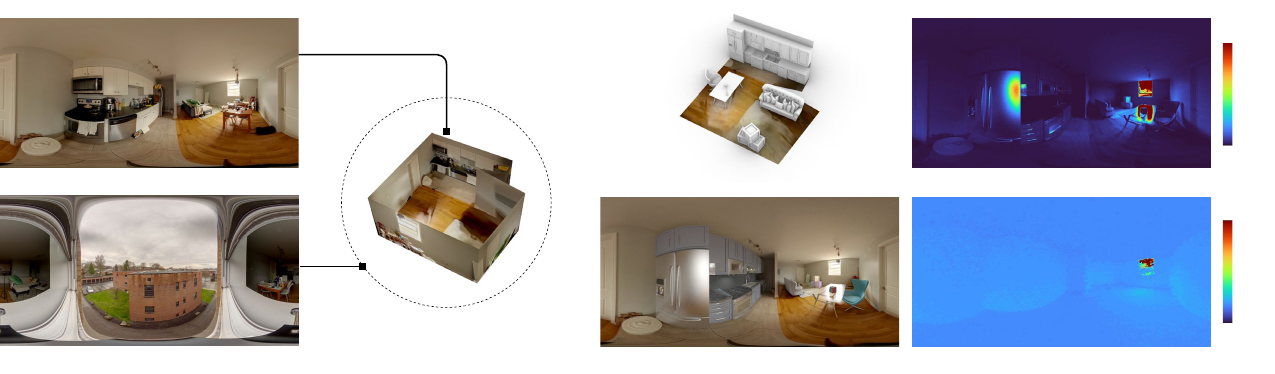}
    \put(4,15){\tiny \color{black}{Indoor Panorama}}
    \put(4,1){\tiny \color{black}{Outdoor Panorama}}
    \put(30,1){\tiny \color{black}{Virtual Model}}
    \put(51,15){\tiny \color{black}{Virtual Objects}}
    \put(53,1){\tiny \color{black}{Virtual Scene}}
    \put(75,15){\tiny \color{black}{Light Map ($cd/m^2$)}}
    \put(96,27){\tiny \color{black}{3,000}}
    \put(96,16){\tiny \color{black}{0}}
    \put(78,1){\tiny \color{black}{Heat Map ($^{\circ}\mathrm{C}$)}}
    \put(96,13){\tiny \color{black}{25}}
    \put(96,2){\tiny \color{black}{21}}    
  \end{overpic}
  \caption{Our pipeline for indoor light and heat estimation: First, a single camera (Ricoh Theta Z1) captures paired indoor and outdoor panoramas. Second, the indoor panorama is used to estimate the 3D floor layout, while the outdoor panorama provides a 360$^{\circ}$ environment map for 3D scene model. Third, virtual objects are inserted into the scene, and the objects are relit within the scene. Finally, an indoor light map with absolute light level ($cd/m^2$) and a heat map ($^{\circ}\mathrm{C}$) are computed for this scene.}
  \label{fig_workflow}
\end{figure*}

In this work, using paired indoor–outdoor HDR panoramas as input, we combine physics-based and data-driven methods to estimate both light and heat properties within indoor environments. The proposed approach supports applications such as virtual home staging and immersive visualization (Fig. \ref{fig_ht_teaser}). By estimating heat from visible-light information, the method quantifies indoor thermal effects contributed by outdoor natural light. In summary, this study makes two primary technical contributions:

\textbf{Visible-Light Estimation} We propose a method that establishes a connection between indoor light transport and heat transfer. Using photometrically calibrated indoor–outdoor HDR panoramas, our application reconstructs radiometrically consistent indoor scenes and estimates spatially varying lighting and thermal properties for panoramic visualization (Fig. \ref{fig_workflow}).

\textbf{Transient Heat Estimation} We implement a transient heat simulation model that quantifies the contribution of solar heat gain to indoor thermal distribution. The model accounts for conduction, convection, and radiation, and can be customized through user-defined thermal parameters. The simulation produces panoramic heat maps that change over time. The estimated heat maps are validated against thermal images captured in real environments.

\section{Related Work}
\subsection{Indoor Light Estimation}
Directly estimating indoor light distribution from a single image is an ill-posed problem. Given a 2D perspective image, some studies focus on predicting panoramic HDR environment maps \cite{gardner2017learning,gkitsas2020deep,legendre2019deeplight,gardner2019deep}, and extensive research efforts have been devoted to estimating HDR from LDR images using inverse tone mapping algorithms \cite{rempel2007ldr2hdr,banterle2006inverse,banterle2007framework,reinhard2002photographic}. The estimated HDR panorama is used as a global light source for high-quality relighting and inserting new virtual 3D objects into the scene. Previous studies have also explored indoor lighting editing \cite{li2022physically}, material property estimation \cite{yeh2022photoscene}, and recovery of spatially-varying lighting \cite{li2020inverse,garon2019fast,srinivasan2020lighthouse} from images. 

Since the input 2D perspective image is in LDR format, and the estimated HDR environment map from previous work is used as a global light source, the resulting scene often fails to reflect accurate real-world radiance. More importantly, earlier studies that estimate environment maps directly from images do not construct a complete light transport model. An ideal model should account for outdoor lighting, scene geometry, and material properties to fully represent the interaction between light and the environment.

\subsection{Outdoor Light Estimation}
Outdoor light varies with sky appearance, surrounding context, and time of day. Luminance and angular distribution of the sky are further influenced by cloud cover, season, and geographic location. Perez sky model \cite{perez1993all} and CIE standard sky model \cite{cie2003spitial} are widely used to estimate outdoor illumination. However, due to subtle differences in cloud cover and sunlight intensity, generalized models cannot fully capture realistic sky appearance \cite{darula2002cie}. Recent studies have aimed to estimate spectral appearance of the sky, which accounts for chromatic variations based on time, sky turbidity, and geographic location \cite{hosek2012analytic}. In addition, analytical models \cite{lalonde2012estimating,preetham1999practical} have been developed to approximate full-spectrum daylight under diverse atmospheric conditions.

Some studies aim to estimate outdoor lighting directly from 2D images. Learning-based methods estimate 360$^{\circ}$ outdoor environment maps from a single 2D photograph \cite{hold2017deep,lalonde2014lighting}, and the resulting HDR sky models are used for relighting tasks. Zhang and Lalonde \cite{zhang2017learning} recover linear HDR outdoor images from a single LDR panorama. Sunkavalli et al. \cite{sunkavalli2008color} reconstruct time-varying color changes in outdoor scenes under direct sunlight and ambient skylight. More recently, Dastjerdi et al. \cite{dastjerdi2023everlight} proposed editing indoor and outdoor illumination sources directly in a panoramic representation. However, these studies primarily focus on estimating outdoor environment maps for relighting virtual objects, without modeling indoor light transport from outdoor sources.

\subsection{Thermal Imaging Technique}
Above absolute zero temperature, indoor objects radiate infrared energy in the Long-Wave Infrared (LWIR) spectrum (wavelengths between 8 $\mu\text{m}$ and 15 $\mu\text{m}$), resulting in heat on the object's surface. The radiometric quantity of the heat transfer can be measured by infrared thermal imaging technique \cite{holst2000common,kaplan2007practical,vollmer2020infrared,planinsic2011infrared}. Thermal imaging has gained increasing attention in computer vision research and leads to applications based on the thermal spectrum, such as object segmentation \cite{huo2023glass}, 3D estimation \cite{eren2009scanning,shin2023deep,narayanan2025shape}, and material property estimation \cite{dashpute2023thermal}. Meanwhile, thermography has been widely used in building analysis, from detecting facade heat loss \cite{motayyeb2023fusion} to integrating thermal data with 3D point clouds \cite{jarzkabek2020supervised}. In indoor settings, it supports 3D modeling \cite{adan2017fusion,schmoll2022method}, fault detection \cite{garrido2018autonomous}, thermal region segmentation \cite{adan2020temporal}, and spatial mapping using thermal and RGB-D cameras \cite{vidas20133d,lopez2017thermographic}.

Existing studies on indoor thermography primarily focus on capturing temporal heat maps at specific time points, but several limitations remain. First, 2D thermography provides only a limited field of view. Second, acquiring multiple 2D perspective images requires additional post-processing, such as image registration and alignment. Third, integrating infrared cameras with devices like RGB-D sensors or LiDAR increases system complexity and data acquisition costs.

\section{Methodology}
\subsection{Indoor Light Estimation}
\label{indoor_lit}
Beyond the RGB image, the absolute light level is crucial for indoor activities. The virtual scene is rendered in HDR format and then converted into luminance values through per-pixel computation~\cite{inanici2006evaluation}. Since both indoor and outdoor HDR images are calibrated to accurately represent real-world scene radiance, this process ensures that the luminance value ($cd/m^2$) in the virtual rendered scene is accurately estimated through physics-based rendering.

In addition to luminance (scene radiance), illuminance (scene irradiance) is commonly used to evaluate the adequacy of indoor task lighting. Each indoor task has its own specific target light level to ensure optimal lighting conditions. To convert and obtain the illuminance map, we assume the scene object is Lambertian surface, where the reflectance ($R$) of the diffuse scene materials are known. The global illuminance map can be derived from displayed luminance value on HDR image \cite{yang2013multi}. 

The conversion of Illuminance ($E$) ($lx$) from Luminance ($L$) ($cd/m^2$) is:
\begin{equation}
    E = \frac{L \pi}{R} 
    \label{eq:E2L_convert}
\end{equation}

\subsection{Indoor Heat Estimation}
\label{indoor_heat}
The heat estimation requires absolute energy influx for each indoor location. We rendered the existing scene in HDR format and then used per-pixel computation to convert it into the corresponding luminance map. Typically, converting illuminance to energy influx requires specialized measurement equipment. Because we focus on the indoor scene under natural light, the per-pixel energy influx can be approximated directly from the illuminance values by using paired HDR images and a physics-based rendering process (obtained from Eq. \ref{eq:E2L_convert}). 
The influx ($W/m^2$) of solar energy is converted from illuminance ($lux$) caused by sunlight. We apply an engineering rule of thumb that 120$lx$ equals 1$W/m^2$ \cite{michael2020conversion}. The per-pixel energy influx will be used as an absolute heat source to compute the indoor heat map.

\subsection{Transient Heat Simulation}
Heat on indoor surfaces varies over time, influenced by material properties (like thermal conductivity, density, and thickness), heat capacity, and surrounding temperatures. We aim to implement a transient heat transport equation to estimate the heat generated by natural light over time.

Thermal diffusivity ($\alpha$) of a material can be expressed as:
\begin{equation}
    \alpha =  \frac{k}{\rho c_p}
    \label{eq:heateq-v1}
\end{equation}

Where $k$ is the thermal conductivity of the material ($W/m\cdot K$), $\rho$ is the density of the material ($kg/m^3$), and $c_p$ is the specific heat capacity of the material at constant pressure ($J/(kg\cdot K)$). The indoor scene has a wide range of building materials, and those variables can be accessed from the existing material database \cite{iesveTableThermal}.

Intrinsic attributes, such as the Laplace operator, are fundamental quantities in various physical simulations, including heat transfer. The work by Narayanan et al. \cite{narayanan2025shape} estimates intrinsic 3D shapes from thermal videos, where transient heat transport follows:

\begin{equation}
    \frac{\mathrm{dT}}{\mathrm{dt}} = \alpha \Delta T + \mathbbm{1}_{\partial \Omega} \frac{1}{\rho c_p \mathrm{dv}} \Bigl(\sigma \epsilon A (T_{surr}^4 - T^4) + h_c A (T_{surr} - T) + A\beta \phi_q \Bigr) 
    \label{eq:heateq-v2}
\end{equation}

Here, $\mathbbm{1}_{\partial \Omega}$ is an indicator function that is 1 if the volume is exposed to the surface and 0 otherwise. The first term $\Delta T$ represents conduction, and the second term represents the sum of radiation, convection, and input heat source;  other quantities are explained in Table \ref{tab:quantities}.

The Eq. \ref{eq:heateq-v2} applies to the object heated by three transfer approaches given the surrounding temperature. For our indoor heat estimation, we provide per-pixel energy influx as the heat source and apply the remaining thermal properties using known values. Besides, for indoor surfaces, heat changes not only by spatially-varying light distribution but also by heat transfer between indoor and outdoor spaces. Therefore, we include heat exchange between indoor and outdoor spaces in the heat equation and add outdoor temperature ($T_{out}$) to the heat transport equation:

\begin{equation}
    \frac{\mathrm{dT}}{\mathrm{dt}} = \alpha \Delta T + \mathbbm{1}_{\partial \Omega} \frac{1}{\rho c_p \mathrm{dv}} \Bigl(\sigma \epsilon A (T_{surr}^4 - T^4) + h_c A (T_{surr} - T) + A\beta \phi_q + A(T - T_{out})\Bigr) 
    \label{eq:heateq-v3}
\end{equation}

\begin{table}[ht]  
    \centering
    \begin{minipage}{0.75\textwidth} 
        \caption{Properties, descriptions, and units for the variables in the Eq. \ref{eq:heateq-v1}, Eq. \ref{eq:heateq-v2}, and Eq. \ref{eq:heateq-v3}.}
        \label{tab:quantities}
        \resizebox{\textwidth}{!}{ 
            \begin{tabular}{@{}c@{\hskip 4mm}c@{\hskip 4mm}c@{}}
                \toprule
                Property & Description & Unit \\
                \midrule 
                $h_c$ & Convection coefficient & $W/(m^2\cdot K)$ \\
                
                $k$ & Thermal conductivity of the material & $W/m\cdot K$ \\
                $\rho$ & Density of the material & $kg/m^3$ \\
                $c_p$ & Specific heat capacity & $J/(kg\cdot K)$ \\
                
                $\epsilon$ & Emissivity & -- \\
                $\beta$ & Energy absorption factor & -- \\
                $\phi_q$ & Input heat flux density & $W/m^2$ \\
                $A$ & Surface area & $m^2$ \\
                
                $\sigma$ & Stefan-Boltzmann constant & $W/(m^2 K^4)$ \\
                $T$ & Temperature & $K$ \\
                $T_{surr}$ & Surrounding temperature & $K$ \\
                $T_{out}$ & Outdoor temperature & $K$ \\
                $\alpha$ & Thermal diffusivity & $m^2/s$ \\
                \bottomrule
            \end{tabular}
        }
    \end{minipage}
\end{table}

\begin{figure*}
  \centering
  \begin{overpic}[width= \textwidth]{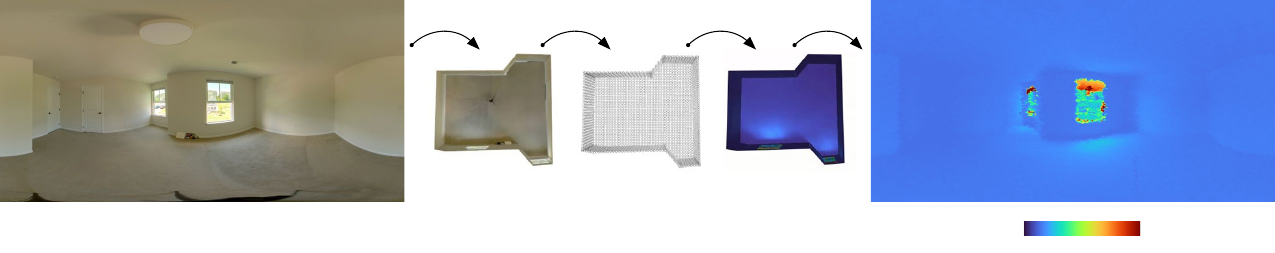}
    \put(0,1.5){\tiny \color{black}{(a)}}
    \put(36,6){\tiny \color{black}{(b)}}
    \put(48,6){\tiny \color{black}{(c)}}
    \put(60,6){\tiny \color{black}{(d)}} 
    \put(68,1.5){\tiny \color{black}{(e)}}
    \put(74,1.5){\tiny \color{black}{294$^\circ\text{F}$}}
    \put(90,1.5){\tiny \color{black}{297$^\circ\text{F}$}} 
  \end{overpic}
  \caption{Heat simulation process: (a) The input is a single panorama; (b) a 3D layout is reconstructed from the input panorama; (c) planar mesh surfaces are uniformly sampled to generate a grid of vertices; (d) the heat equation is applied to compute the per-pixel heat value at each vertex; and (e) the output heat map is projected from 3D coordinates onto an equirectangular representation.}
  \label{fig_ht_simulation}
\end{figure*}

We estimate the indoor heat map using the transient heat equation (Eq. \ref{eq:heateq-v3}) to calculate the temperature at each pixel. In Fig. \ref{fig_ht_simulation}, the 3D floor layout is derived from a single indoor panorama. The heat simulation is conducted in 3D coordinates, with planar surfaces subdivided into vertices to compute the heat value for each point. After the simulation, the heat data in 3D coordinates is mapped onto an equirectangular representation and displayed from the same camera position as the input panorama.

\begin{figure*}[t]
  \centering
  \begin{overpic}[width= \textwidth]{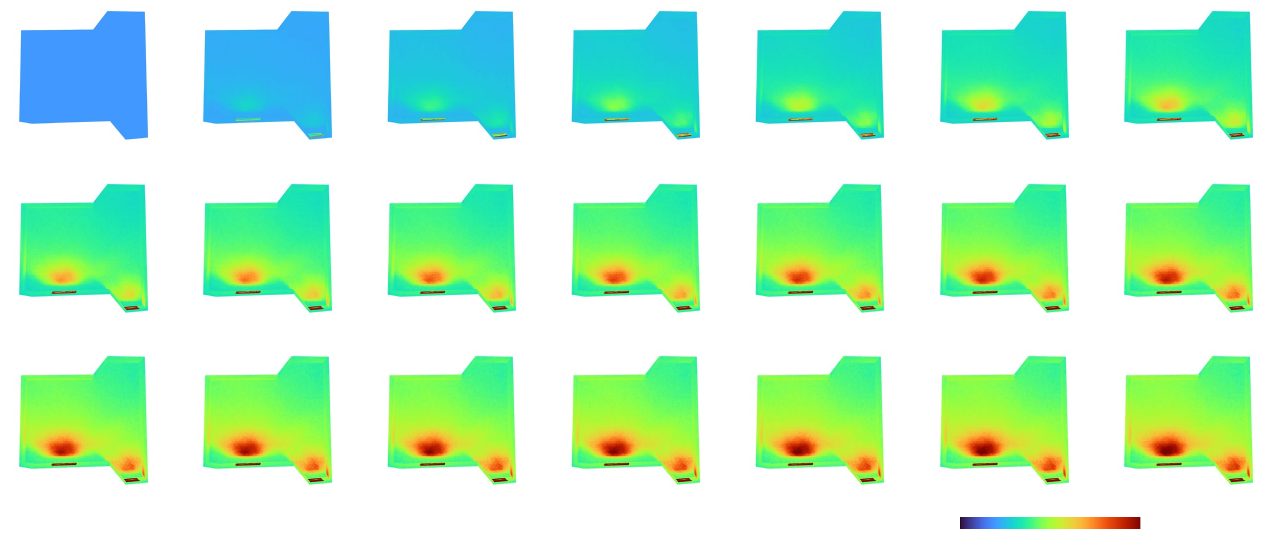}
    \put(2,31){\tiny \color{black}{0s}}
    \put(16,31){\tiny \color{black}{10s}}
    \put(30,31){\tiny \color{black}{20s}}
    \put(44,31){\tiny \color{black}{30s}}
    \put(60,31){\tiny \color{black}{40s}}
    \put(74,31){\tiny \color{black}{50s}}
    \put(88,31){\tiny \color{black}{60s}}

    \put(2,17.5){\tiny \color{black}{70s}}
    \put(16,17.5){\tiny \color{black}{80s}}
    \put(30,17.5){\tiny \color{black}{90s}}
    \put(44,17.5){\tiny \color{black}{100s}}
    \put(60,17.5){\tiny \color{black}{110s}}
    \put(74,17.5){\tiny \color{black}{120s}}
    \put(88,17.5){\tiny \color{black}{130s}}

    \put(2,4){\tiny \color{black}{140s}}
    \put(16,4){\tiny \color{black}{150s}}
    \put(30,4){\tiny \color{black}{160s}}
    \put(44,4){\tiny \color{black}{170s}}
    \put(60,4){\tiny \color{black}{180s}}
    \put(74,4){\tiny \color{black}{190s}}
    \put(88,4){\tiny \color{black}{200s}}
    
    \put(69,1.5){\tiny \color{black}{294$^\circ\text{F}$}}
    \put(90,1.5){\tiny \color{black}{297$^\circ\text{F}$}}   
  \end{overpic}
  \caption{Transient heat simulation over time: The heat map is initialized and recorded at successive time steps, shown from a top-down view to illustrate the 3D floor layout as heat propagates from 0$s$ to 200$s$.}
  \label{fig_transient_ht_step}
\end{figure*}

The transient heat simulation allows the heat values to be visualized over time. In Fig. \ref{fig_transient_ht_step}, we selected a scene to estimate heat changes on the 3D layout from a top-down perspective. The results of the heat simulation are recorded at time steps, where the 0$s$ figure shows the indoor space at the initial temperature. As time progresses, the indoor space heats up, and the indoor temperature is captured incrementally at each step.

\begin{figure*}[t]
  \centering
  \begin{overpic}[width= \textwidth]{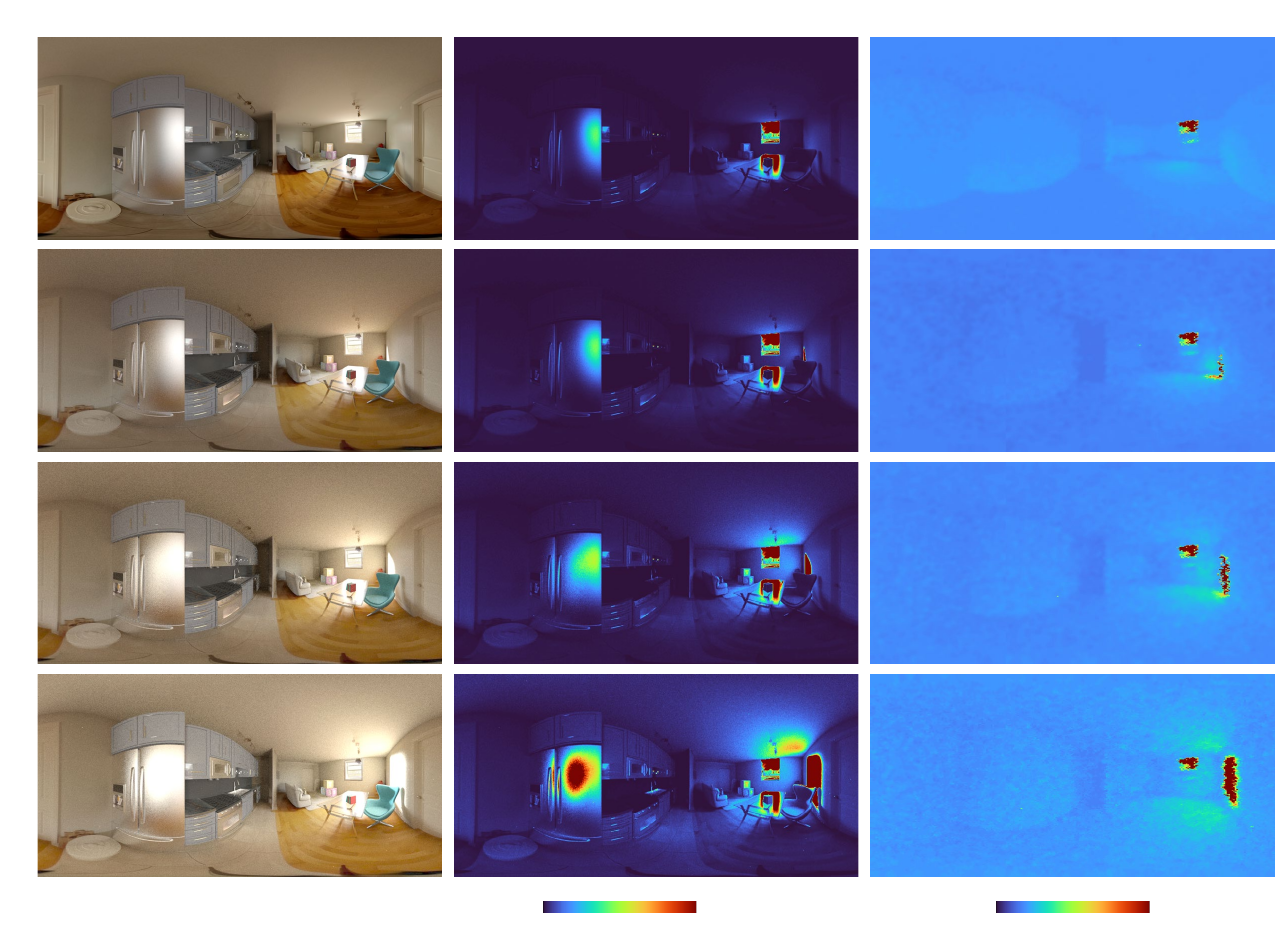}
    \put(10,71){\tiny \color{black}{Rendered Scene}}
    \put(45,71){\tiny \color{black}{Light Map}}
    \put(78,71){\tiny \color{black}{Heat Map}}
    \put(0,62){\tiny \color{black}{(a)}}
    \put(0,46){\tiny \color{black}{(b)}}
    \put(0,28){\tiny \color{black}{(c)}}
    \put(0,12){\tiny \color{black}{(d)}}   
    \put(41,1.5){\tiny \color{black}{0}}
    \put(55,1.5){\tiny \color{black}{500 $cd/m^2$}}
    \put(73,1.5){\tiny \color{black}{21$^\circ\text{C}$}}
    \put(91,1.5){\tiny \color{black}{25$^\circ\text{C}$}} 
  \end{overpic}
  \caption{Indoor light and heat estimation under new outdoor sun illuminations: The virtual scene is rendered under the captured outdoor panorama (a). The same scene is then rendered with a virtual sun at 16:45 (b), 17:45 (c), and 18:45 (d), respectively.}
  \label{fig_ht_change_sun}
\end{figure*}

\begin{figure}[t]
  \centering
  \begin{overpic}[width= \textwidth]{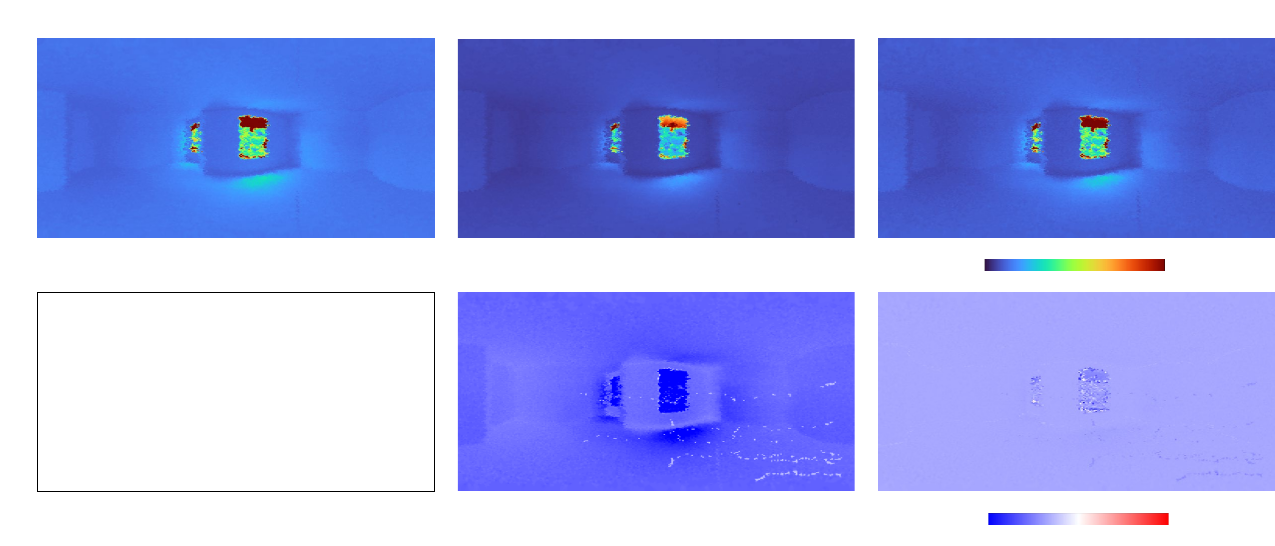}
    \put(12,41){\tiny \color{black}{Plaster (Dense)}}
    \put(48,41){\tiny \color{black}{Glass}}
    \put(82,41){\tiny \color{black}{Wood}}
    \put(1,32){\tiny{\makebox(0,0){\rotatebox{90}{Heat Map}}}}
    \put(1,12){\tiny{\makebox(0,0){\rotatebox{90}{Difference Map}}}}
    \put(72,22){\tiny \color{black}{21$^\circ\text{C}$}}
    \put(92,22){\tiny \color{black}{26$^\circ\text{C}$}} 
    \put(72,2){\tiny \color{black}{-2$^\circ\text{C}$}}
    \put(93,2){\tiny \color{black}{2$^\circ\text{C}$}}
  \end{overpic}
  \caption{Heat estimation across different surface materials: The heat map row illustrates the estimated heat panoramas generated for each material. The error map row uses the estimated heat map to subtract the baseline heat map (under plaster (dense)) and displays the per-pixel temperature error ($^\circ\text{C}$).}
  \label{fig_material_error}
\end{figure}

Our application can estimate indoor light and heat panoramas based on new outdoor images. In Fig. \ref{fig_ht_change_sun}, we present the joint estimation of the indoor light and heat properties for the scene. The existing scene is renovated with new furniture objects and then converted into the corresponding light and heat maps. Following the approach of Ji et al. \cite{ji2024virtual}, the outdoor scene is initially cloudy, and direct sunlight is added to the outdoor image and used as a light source to relight the scene at different times in the afternoon (16:45, 17:45 and 18:45). Compared to the cloudy scenario, direct sunlight increases indoor light levels on the table and walls, leading strong reflections on the refrigerator, while also increasing temperatures on wall surfaces near the window.

The transient heat simulation can be performed with different surface materials \cite{iesveTableThermal}. As shown in Fig. \ref{fig_material_error}, we first used plaster (dense) as the surface material for the scene, and the resulting estimated heat map served as the baseline. We then replaced the surface material with glass and wood, generating the corresponding heat maps for the same scene. Compared with plaster (dense), both glass and wood produced lower surface temperatures.

\begin{figure}[t]
  \centering
  \begin{overpic}[width= \textwidth]{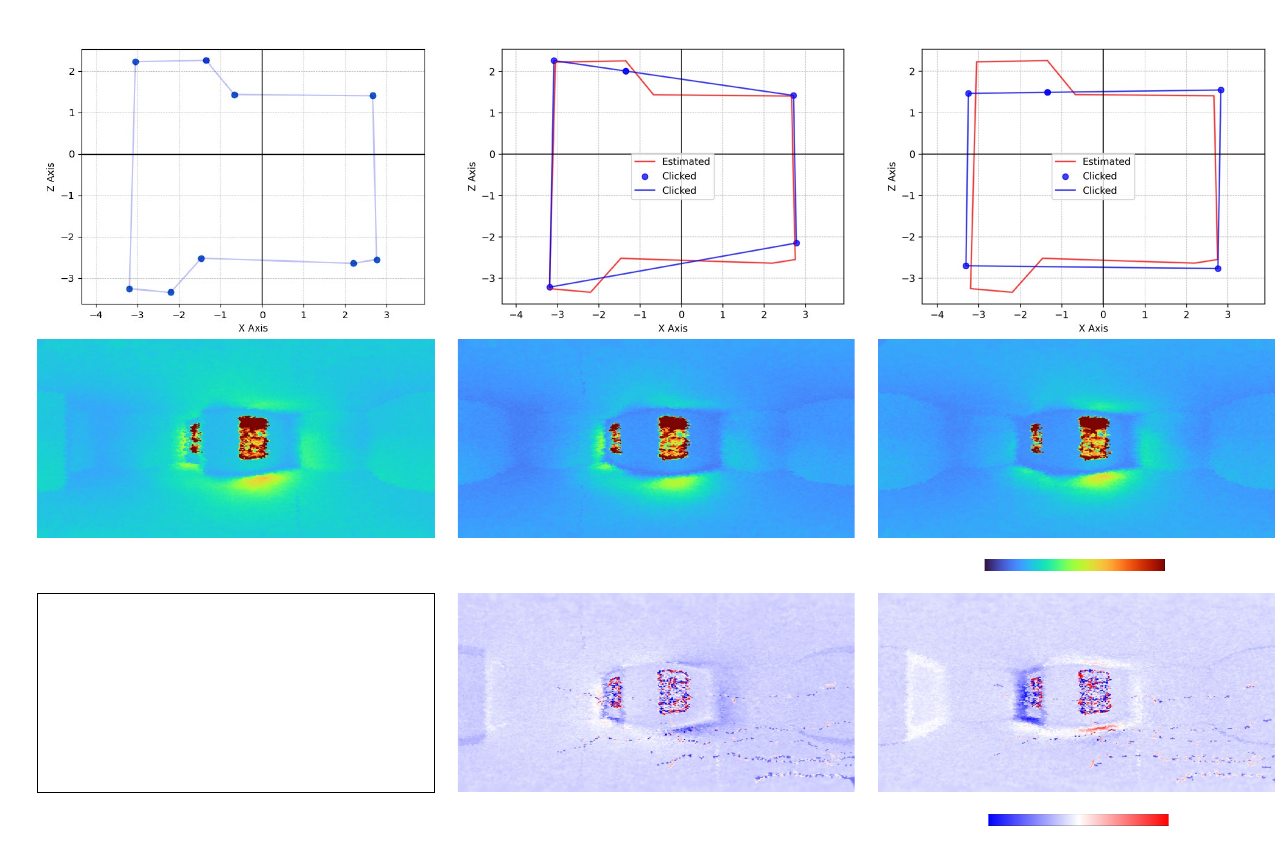}
    \put(14,64){\tiny \color{black}{Layout 1}}
    \put(47,64){\tiny \color{black}{Layout 2}}
    \put(80,64){\tiny \color{black}{Layout 3}}
    \put(1,54){\tiny{\makebox(0,0){\rotatebox{90}{Floor Boundary}}}}
    \put(1,30){\tiny{\makebox(0,0){\rotatebox{90}{Heat Map}}}}
    \put(1,12){\tiny{\makebox(0,0){\rotatebox{90}{Error Map}}}}
    \put(72,22){\tiny \color{black}{21$^\circ\text{C}$}}
    \put(92,22){\tiny \color{black}{26.0$^\circ\text{C}$}} 
    \put(72,1.5){\tiny \color{black}{-2$^\circ\text{C}$}}
    \put(93,1.5){\tiny \color{black}{2$^\circ\text{C}$}}     
  \end{overpic}
  \caption{Error analysis of heat estimation across floor layouts: Layout 1 shows the correct floor boundary, while Layout 2 and Layout 3 modify it. The heat map illustrates the estimated heat panoramas. The error map uses the estimated heat map to minimize the baseline heat map (from Layout 1).}
  \label{fig_ht_geo_error}
\end{figure}

An accurate 3D floor layout is essential for precise heat estimation. We focus on the error analysis of the heat simulation in different floor layouts (Fig. \ref{fig_ht_geo_error}). Layout 1 is considered as the correct floor layout, while Layouts 2 and 3 feature slight shifts in the floor boundaries. The heat maps for Layouts 2 and 3 reveal variations in indoor temperature values due to inaccuracies in the room geometry. In the error map, the heat maps of Layouts 2 and 3 are compared with the ground-truth (GT) heat map of Layout 1. Layout 2 shows lower temperature estimates at the upper and lower corners of the wall, while Layout 3 exhibits lower temperatures near the wall adjacent to the left window. The 3D layout estimation assumes a Manhattan layout, which is effective for regular indoor geometries. Curved walls and tilted roofs are simplified into a cubic model.

\begin{figure*}[t]
  \centering
  \begin{overpic}[width= \textwidth]{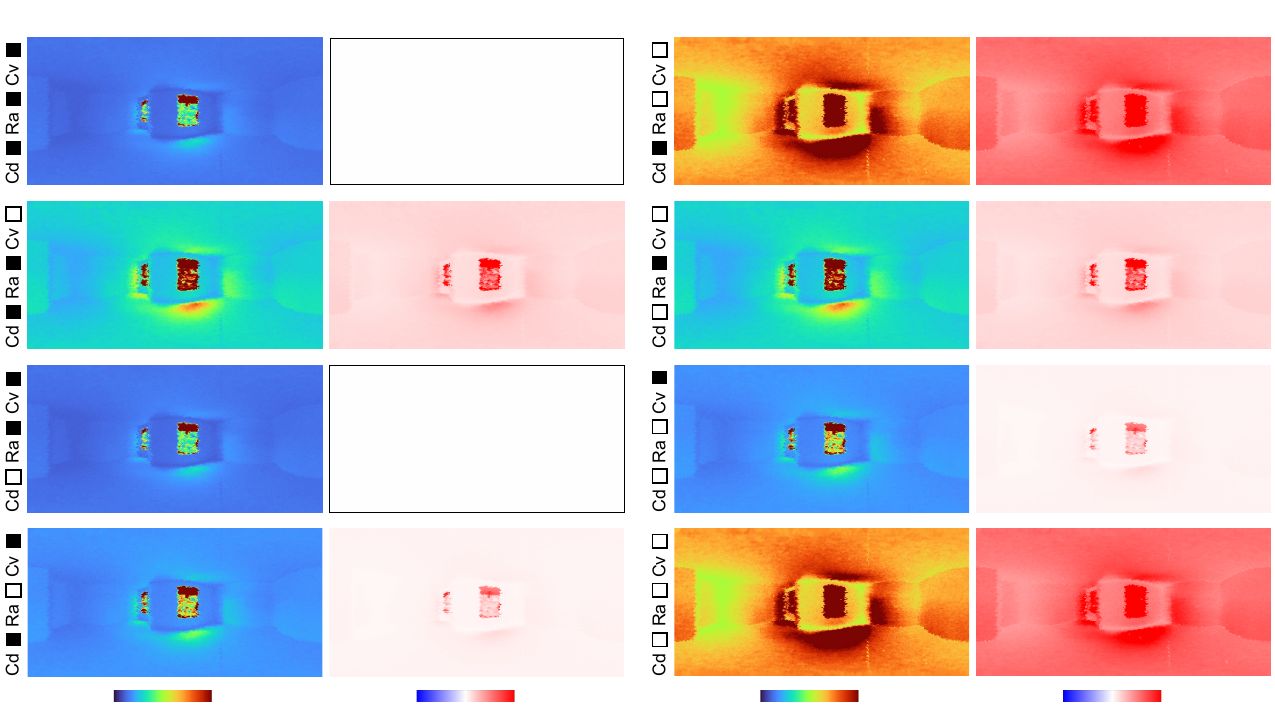}
    \put(9,54){\tiny \color{black}{Heat Map}} 
    \put(32,54){\tiny \color{black}{Error Map}}     
    \put(59,54){\tiny \color{black}{Heat Map}} 
    \put(82,54){\tiny \color{black}{Error Map}}   
    \put(4,1){\tiny \color{black}{21$^\circ\text{C}$}}
    \put(17,1){\tiny \color{black}{31$^\circ\text{C}$}}
    \put(27,1){\tiny \color{black}{-10$^\circ\text{C}$}}
    \put(41,1){\tiny \color{black}{10$^\circ\text{C}$}} 
    \put(54,1){\tiny \color{black}{21$^\circ\text{C}$}}
    \put(68,1){\tiny \color{black}{31$^\circ\text{C}$}}
    \put(77,1){\tiny \color{black}{-10$^\circ\text{C}$}}
    \put(92,1){\tiny \color{black}{10$^\circ\text{C}$}}     
  \end{overpic}
  \caption{Effects of three heat-transfer types in heat simulations (Eq. \ref{eq:heateq-v2}): Cd (conduction), Ra (radiation), and Cv (convection). Black squares indicate when the corresponding heat-transfer type is included in the transient heat simulation. Heat map displays results for each combination, while error map compares each case to the baseline (all components included). Because of the low material conductivity (plaster: 0.5 $W/(m\cdot K)$), conduction has minimal effect, producing only minor differences in the third-row, left-column error map.}
  \label{fig_3ht_transf}
\end{figure*}

\subsection{Sensitivity Analysis}
The transient heat simulation (Eq. \ref{eq:heateq-v2}) accounts for three types of heat transfer: conduction, radiation, and convection. To evaluate the impact of each component on heat estimation, parametric heat simulations were performed with various combinations of these components, and the corresponding heat maps for each condition are visualized, as shown in Fig. \ref{fig_3ht_transf}. When conduction, radiation, and convection are all included, the resulting heat map is considered the baseline for computing the error map. When there is no heat transfer (such as conduction, radiation, and convection), the temperature remains at its highest, as no heat is emitted from the object's surface. Adding different types of heat transfer results in varying reductions in surface temperature. 
Each heat map is compared with the baseline image (first row, left column). In the error maps, the baseline image displays a matrix of zero values, indicating the absence of temperature differences. Owing to the low material conductivity in the heat simulation, the image in the third row, left column exhibits minimal temperature variation. Other combinations of heat transfer mechanisms increase the temperature to varying degrees, highlighting their individual and collective contributions to the overall heat distribution.

\begin{figure*}[t]
  \centering
  \begin{overpic}[width= \textwidth]{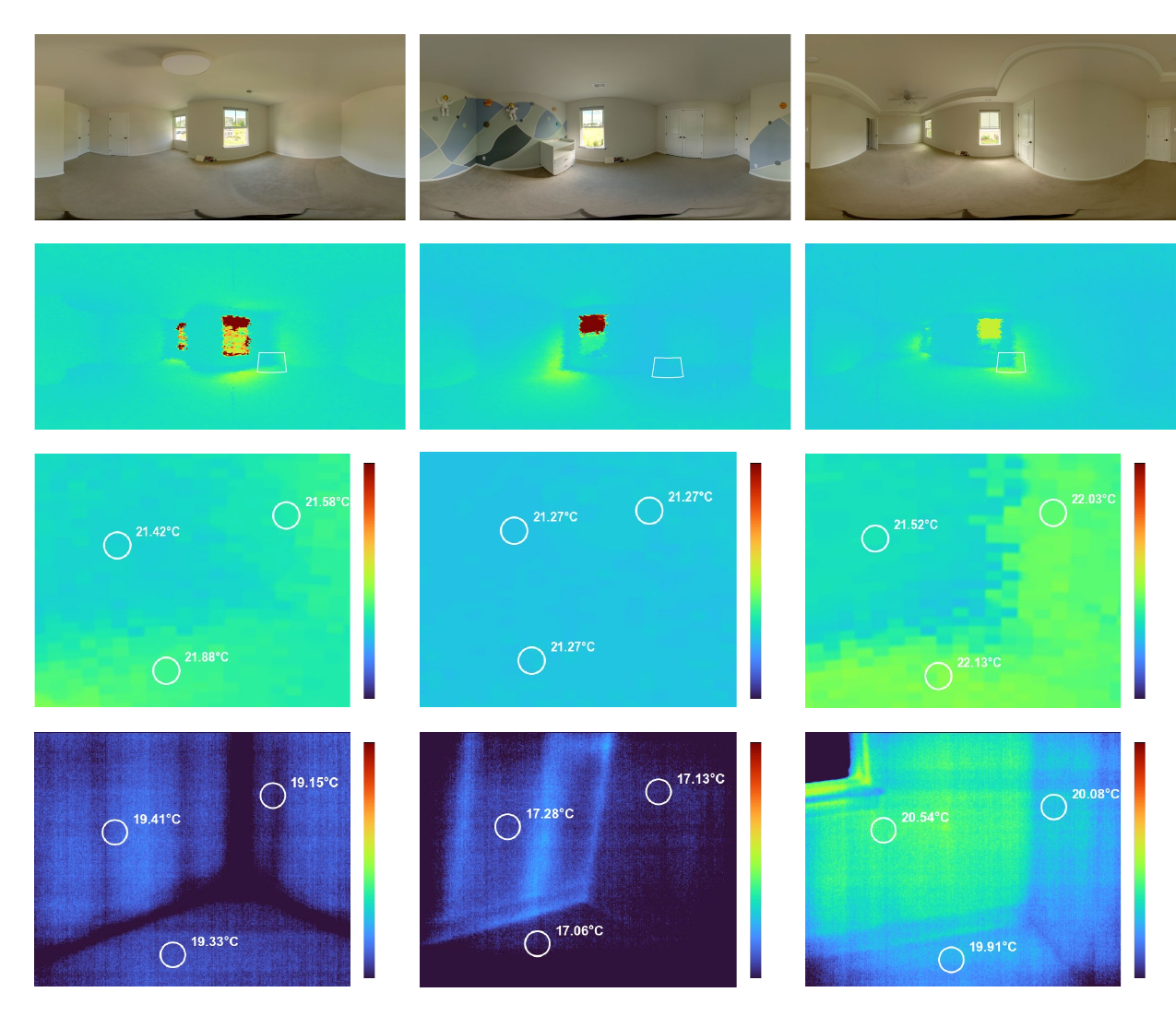}
    \put(15,84){\scriptsize \color{black}{Scene 1}}
    \put(47,84){\scriptsize \color{black}{Scene 2}}
    \put(79,84){\scriptsize \color{black}{Scene 3}}
    \put(1,75){\scriptsize{\makebox(0,0){\rotatebox{90}{Input}}}}  
    \put(1,57){\scriptsize{\makebox(0,0){\rotatebox{90}{Heat Map}}}} 
    \put(1,37){\scriptsize{\makebox(0,0){\rotatebox{90}{2D Perspective}}}} 
    \put(1,13){\scriptsize{\makebox(0,0){\rotatebox{90}{Thermal Camera}}}}  
    
    \put(33,28){\tiny{\makebox(0,0){\rotatebox{90}{20.0}}}}     
    \put(33,46){\tiny{\makebox(0,0){\rotatebox{90}{25.0}}}}  
    \put(66,28){\tiny{\makebox(0,0){\rotatebox{90}{20.0}}}}     
    \put(66,46){\tiny{\makebox(0,0){\rotatebox{90}{25.0}}}}
    \put(99,28){\tiny{\makebox(0,0){\rotatebox{90}{20.0}}}}     
    \put(99,46){\tiny{\makebox(0,0){\rotatebox{90}{25.0}}}} 

    \put(33,4){\tiny{\makebox(0,0){\rotatebox{90}{19.0}}}}     
    \put(33,22){\tiny{\makebox(0,0){\rotatebox{90}{23.0}}}}  
    \put(66,4){\tiny{\makebox(0,0){\rotatebox{90}{17.0}}}}     
    \put(66,22){\tiny{\makebox(0,0){\rotatebox{90}{21.0}}}}
    \put(99,4){\tiny{\makebox(0,0){\rotatebox{90}{19.0}}}}     
    \put(99,22){\tiny{\makebox(0,0){\rotatebox{90}{23.0}}}}   
    
    \put(81,0){\scriptsize \color{black}{Temperature ($^\circ\text{C}$)}}
  \end{overpic}
  \caption{Comparison between the rendered heat map and thermal images captured on-site: Three indoor scenes were captured as inputs, and panoramic heat maps were generated through simulation. The target regions were cropped to align with the thermal camera’s field of view and pose. Indoor surface temperatures are highlighted in white, and the temperature scale of each heat map is adjusted for visualization clarity.}
  \label{fig_ht_comp}
\end{figure*}

\subsection{Validation and Comparison}
To evaluate the results of our heat simulation, we captured real-world scenes using a Ricoh Theta Z1 camera alongside a thermal camera. The heat simulation enables visualization of indoor heat maps in the equirectangular representation, while the thermal camera provides only a 2D perspective view. Therefore, we approximately cropped the target region from the rendered heat map and converted it into a 2D perspective with the same field of view (FOV) as the thermal camera. We then evaluated the rendered images by comparing them with the thermal images captured by the thermal camera.

When capturing visible-light and thermal images, a Ricoh Theta Z1 camera is positioned side by side with the FLIR thermal camera at the same height, with a Macbeth color checker and a gray matte board placed within the field of view. The real-time indoor temperature is recorded on the Govee Hygrometer Thermometer (H5075), and the outdoor temperature is obtained from GPS data and local time.

We set up the FLIR thermal camera in TLinear mode (Boson, image resolution: 640 by 512 pixels) when capturing the thermal images. However, TLinear mode assumes the default emissivity, atmosphere temperature, and background temperature for the scene. The thermal image captured by TLinear requires a post-process \cite{flirpyThermalCameras} to display the correct thermal image based on the actual emissivity of the materials.

Using our proposed heat estimation method, a single indoor panorama is used to generate a panoramic heat map from the same camera position (Fig. \ref{fig_ht_comp}). The average temperature values are calculated on the planar indoor surfaces within the white circles, and the heat estimation results are compared with the thermal images captured by the thermal camera. In Scene 1, the temperature difference in selected areas ranged from around 2.00 to 2.50 $^\circ\text{C}$. In Scene 2, the temperature difference was approximately 4.00 to 4.50 $^\circ\text{C}$. In Scene 3, the wall plane in the upper left corner showed a temperature difference of about 1.00 $^\circ\text{C}$, while the wall plane on the right and the floor exhibited larger temperature errors, close to 2.00 $^\circ\text{C}$. The simulation discrepancy arises from several factors, such as complex heat sources (air conditioning, electronic appliances), uncertainties in material properties, and the simplified Lambertian surface assumption used to model the real-world scene. It is also important to note that the thermal images from the thermal camera are usually noisy, which can lower the average temperature in the target region.

\section{Conclusion}
In this work, we present an application that uses paired indoor and outdoor RGB panoramas as input to estimate indoor lighting and heat properties. Built on image-based rendering techniques, our inverse rendering pipeline not only provides high-quality indoor renderings in a photorealistic manner but also enables us to estimate light levels for indoor task lighting analysis. Since the panorama provides an omnidirectional view of the scene, our application allows users to identify suitable work areas. In addition, heat estimation is integrated with light estimation. The heat generated by natural light on indoor surfaces is estimated, whereas transient heat transport enables the analysis of indoor heat distribution over time. Our method provides an efficient solution for indoor lighting and thermal analysis, supporting virtual indoor staging and contributing to high-quality scene visualization, lighting research, and thermal imaging techniques for real-world scenarios.

\section{Discussion}
\subsection{Limitations}
\label{sec:limit_ht}
This paper presents the methodology for a joint estimation of light and heat for a single visible-light panorama. However, several limitations are identified in this work. The heat estimation assumes that all indoor heat originates from natural light (the outdoor RGB panorama) and takes sunlight as the sole heat source. Since mechanical systems, such as air conditioning or heating operations, are unknown in a single visible-light panorama, their power and functionality remain unknown. Consequently, the heat estimation reflects only the temperature effects caused by outdoor light. The outdoor scene is captured as a 360$^\circ$ environment map. This map serves as an omnidirectional illumination source from an infinite distance. However, accurately modeling detailed outdoor 3D objects can be challenging when obstructions are present. For example, under clear sky conditions, sunlight can be blocked by trees and leaves, resulting in light leakage as beams pass through the foliage. These subtle lighting effects cannot be recovered effectively, as the scene is flattened onto the environment map. The estimation of irradiance (illuminance) is based on the assumption that all objects have Lambertian surfaces. Materials such as glass or metal have not been included in this work.

\subsection{Future Work}
Future work should consider more complex materials, such as transparent glass and specular metals, to enhance the realism of virtual rendered scenes. Additionally, heat simulations should investigate a broader range of material properties and analyze the impacts of various parameters on heat change. Expanding the scope to include more building cases, such as those in different geographic locations, with diverse indoor materials, and under varying weather conditions, would further enrich the study and its applicability.

\section*{Acknowledgement} This work was partially supported by a gift from Zillow Group, USA, and NSF grant IIS2107236.

%
\bibliographystyle{splncs04}
\bibliography{references}




\end{document}